\documentclass[letterpaper]{article}
\usepackage[preprint]{aaai2027}

\usepackage[hyphens]{url}
\usepackage{graphicx}
\usepackage{natbib}
\usepackage{caption}
\usepackage{algorithm}
\usepackage{algorithmic}
\usepackage{amsmath}
\usepackage{amsfonts}
\usepackage{amssymb}

\usepackage{newfloat}
\usepackage{listings}

\DeclareCaptionStyle{ruled}{
    labelfont=normalfont,
    labelsep=colon,
    strut=off
}

\floatstyle{ruled}
\newfloat{listing}{tb}{lst}{}
\floatname{listing}{Listing}

\usepackage{booktabs}

\usepackage{xcolor}

\title{SolarFlowRefiner: Refinement-Aware Flow Matching \\for Surface Solar Radiation Downscaling}

\author{
    Udbhav Srivastava\textsuperscript{\rm 1},
    Antonita Racheal\textsuperscript{\rm 1},
    Yiheng Chen\textsuperscript{\rm 1},
    Runlong Yu\textsuperscript{\rm 1}\corresponding,
    Xinyue Ye\textsuperscript{\rm 2}\corresponding
}

\affiliations{
    \textsuperscript{\rm 1}Department of Computer Science, The University of Alabama, Tuscaloosa, Alabama, USA\\
    \textsuperscript{\rm 2}Department of Geography, The University of Alabama, Tuscaloosa, Alabama, USA\\
    usrivastava@crimson.ua.edu, aracheal@crimson.ua.edu, ychen226@crimson.ua.edu, ryu5@ua.edu, xye10@ua.edu
}

\begin{document}

\maketitle

\begin{abstract}
High-resolution surface solar radiation (SSR) is important for solar forecasting and grid operation. However, physically consistent reanalysis products are too coarse to resolve localized cloud-driven variability. In this paper, we study a multisource downscaling task that reconstructs high-resolution SolarCube SSR fields from coarse ERA5 radiative variables and co-registered satellite channels. 
The task is challenging because a single ERA5 grid cell may contain both sunlit and cloud-shadowed regions. As a result, the missing high-resolution correction can be spatially sharp and inherently ambiguous. One-stage predictors often oversmooth these structures. Post-hoc refinement also introduces a stage-wise mismatch: the generator is optimized independently, even though its output determines the refiner's initial state.
We introduce SolarFlowRefiner, a refinement-aware flow-matching framework for SSR downscaling. A conditional FlowMatch generator first predicts a normalized correction to an upsampled ERA5 baseline. The refiner is then trained on prediction-conditioned states between the current FlowMatch output and the target residual. This exposes the refiner to the structured errors produced by the generator. The refinement objective is also backpropagated through the FlowMatch sampler, allowing generation and correction to be jointly optimized for the final reconstruction.
Experiments on a day-blocked ERA5--SolarCube benchmark show consistent improvements over standalone generation and post-hoc refinement. More broadly, SolarFlowRefiner provides a general strategy for coupling generative predictors with iterative correctors.
\end{abstract}

\section{Introduction}
Solar radiation reaching the surface is quantified as downward shortwave
irradiance \citep{solarhandbook}. This surface flux, composed of direct-beam and
diffuse sky components and strongly modulated by clouds and aerosols, is the
primary driver of photovoltaic generation \citep{pvforecast}. 
Ground pyranometers provide accurate point measurements but sparse spatial
coverage; satellite retrievals provide spatially continuous estimates but rely on
indirect observations; and atmospheric reanalysis provides physically consistent,
gap-free fields at coarse spatial resolution \citep{pvforecast,solarcube}. In this
work we use two complementary sources at opposite ends of this resolution
spectrum. ERA5 provides model-derived radiative fields, including surface downward
shortwave radiation, on a coarse global grid \citep{era5}. SolarCube provides
high-resolution surface solar radiation retrieved from geostationary satellite
observations, together with visible, infrared, and cloud-related channels that
resolve fine-scale cloud structure \citep{goes,solarcube}. Our task links these
sources: recovering the fine-scale surface solar radiation (SSR) field observed by
SolarCube from coarse ERA5 radiative inputs.
 
Accurate, spatially detailed radiation fields are increasingly important as solar
generation is integrated into electricity grids at scale, because the combined
output of distributed solar installations depends on radiation variations at
spatial scales that coarse products cannot resolve \citep{pvforecast}. Passing
clouds create sharp, localized swings in surface irradiance, and these fine-scale
fluctuations drive the sudden ramps in power output that are difficult for grid
operators to balance. Yet the gap-free, physically consistent estimates offered by
reanalysis come at a coarse resolution, roughly $0.25^{\circ}$ for ERA5
\citep{era5}, at which a single grid cell averages over both sunlit and
cloud-shadowed ground. The fine-scale structure that governs local solar
variability is therefore lost. Reconstructing high-resolution radiation fields
from coarse inputs, i.e., spatial super-resolution, is an important problem for
solar forecasting, grid management, and climate applications
\citep{stengel,solarcube}.
 
Surface solar radiation is dominated by cloudcover, whose spatial structure is sharp, intermittent, and fast-moving, and only weakly constrained by coarse reanalysis fields \citep{solarcube,goes}. Whereas clear-sky irradiance varies smoothly with solar geometry, cloud cover imposes abrupt, high-contrast gradients that shift on short timescales. A single coarse ERA5 cell aggregates many distinct cloud and clear-sky conditions, so many high-resolution radiation fields may be consistent with the same coarse observation. This one-to-many ambiguity makes solar super-resolution a severely ill-posed inverse problem. It also explains a common limitation of deterministic models: pixel-wise training tends to regress toward a conditional mean over plausible solutions, producing over-smoothed fields that miss sharp radiation gradients near cloud edges \citep{blaumichaeli,efdiff}. Recovering this structure requires a model that can represent residual uncertainty and a conditioning signal that carries the missing cloud information.
 
Recent super-resolution and generative restoration methods provide the starting
point for this work. Deep convolutional and transformer models have become strong
deterministic predictors for natural-image restoration
\citep{edsr,rcan,swinir,hat}, while generative methods such as residual-shift \cite{resshift} models reconstruct high-resolution images through iterative correction in a structured residual space. More recently, flow matching \citep{flowmatch} has emerged as an efficient, simulation-free generative
formulation, and iterative refinement methods \citep{pderefiner,flowrefiner}
improve an initial prediction through repeated correction rather than regenerating
it from scratch; together these motivate the generation-and-refinement approach we
adopt. However, using generation and refinement as separately trained
stages introduces a mismatch: the generator is optimized to produce the best
standalone residual estimate, even though this estimate later serves as the
initial state for a multi-step correction process. The best standalone prediction
is therefore not necessarily the best starting point for refinement. This
motivates refinement-aware generation, where the generator is trained not only to
be accurate, but also to produce residual states that are easy for the refiner to
improve.
 
We propose SolarFlowRefiner, a refinement-aware FlowMatch framework for
ERA5-to-SolarCube SSR downscaling. FlowMatch first generates an initial normalized
residual conditioned on ERA5 radiative variables and SolarCube auxiliary channels.
A refiner then iteratively corrects the residual. Our main model, SolarFlowRefiner, trains the FlowMatch generator and PDE-style refiner end to end by backpropagating the refinement loss through the FlowMatch sampler. In this way, the generator learns to be refinable. We compare post-hoc refinement variants that keep the base generator fixed: FlowRefiner-ODE learns a continuous correction trajectory, while FlowRefiner-PDE uses the same multilevel denoising refiner as the proposed model but blocks gradients into the FlowMatch generator.

% We organize the paper around the following research questions:
% \begin{itemize}
%     \item \textbf{RQ1:} How effective are deterministic, residual-shift, and
%     flow-based models for ERA5-to-SolarCube SSR residual generation?
%     \item \textbf{RQ2:} Does iterative refinement improve FlowMatch residual
%     generation, and how do ODE-style and PDE-style post-hoc refinement compare?
%     \item \textbf{RQ3:} Does coupling the generator and refiner end to end improve
%     over post-hoc refinement by making the generator refinement-aware?
%     \item \textbf{RQ4:} For SolarFlowRefiner, how much do ERA5 radiative channels
%     and SolarCube satellite channels each contribute to SSR downscaling?
% \end{itemize}

Our contributions are summarized as follows:
\begin{itemize}
    \item We formulate multisource SSR downscaling as residual reconstruction around a physically meaningful ERA5 baseline, focusing learning on fine-scale cloud-driven corrections.

    \item We introduce a prediction-conditioned multilevel refinement scheme that trains the corrector on structured errors produced by the current generator.

    \item We develop a refinement-aware end-to-end objective that jointly optimizes generation and correction for the final SSR reconstruction, and validate its effectiveness against standalone and post-hoc refinement baselines.
\end{itemize}

\section{Related Work}
\label{sec:related_work}

We review geophysical super-resolution and generative refinement, highlighting the gap between post-hoc correction and jointly optimized generation--refinement.

\subsection{Geophysical Super-Resolution}

Deep super-resolution has progressed from convolutional models such as SRCNN and VDSR \citep{srcnn,vdsr} to residual and attention-based architectures including EDSR, RCAN, SwinIR, and HAT \citep{edsr,rcan,swinir,hat}. Adversarial methods such as SRGAN and ESRGAN further target perceptual realism, while exposing the perception--distortion tradeoff \citep{srgan,esgran,blaumichaeli}. These methods provide strong baselines, but their one-stage predictions can oversmooth localized structures when multiple fine-scale fields are compatible with the same coarse input.

Learning-based downscaling has also been applied to climatological wind and solar fields \citep{stengel}, with WiSoSuper providing a benchmark for renewable-energy super-resolution \citep{wisosuper}. SolarCube supplies co-registered satellite observations and high-resolution SSR across multiple sites \citep{solarcube}. We use ERA5 as a coarse physical baseline and SolarCube channels as cloud-scale context, predicting and refining the normalized correction to the ERA5 field rather than directly regressing the full SSR field.

\subsection{Generative Refinement}

Generative super-resolution represents the ambiguity of recovering fine-scale fields from coarse observations. ResShift constructs a residual-shifting process between degraded and high-resolution images \citep{resshift}, while normalizing-flow and rectified-flow methods learn conditional transport paths for restoration \citep{srflow,flowmatch,rectifiedflow,flowsr,onestepsr,flowie,pmrf}. These approaches motivate our use of FlowMatch to generate an initial cloud-conditioned SSR residual.

Iterative methods instead improve an existing prediction. PDE-Refiner uses multilevel denoising to recover spatial-frequency content missed by one-shot predictors \citep{pderefiner}, whereas FlowRefiner learns a continuous ODE-style correction trajectory toward the target \citep{flowrefiner}. Both naturally support post-hoc correction of a fixed prediction. SolarFlowRefiner differs by constructing refinement states from the current FlowMatch output and backpropagating the refinement objective through the sampler, thereby optimizing the generator and refiner jointly for the final SSR reconstruction rather than as independent stages.

\begin{figure*}[t]
\centering
\includegraphics[width=\textwidth]{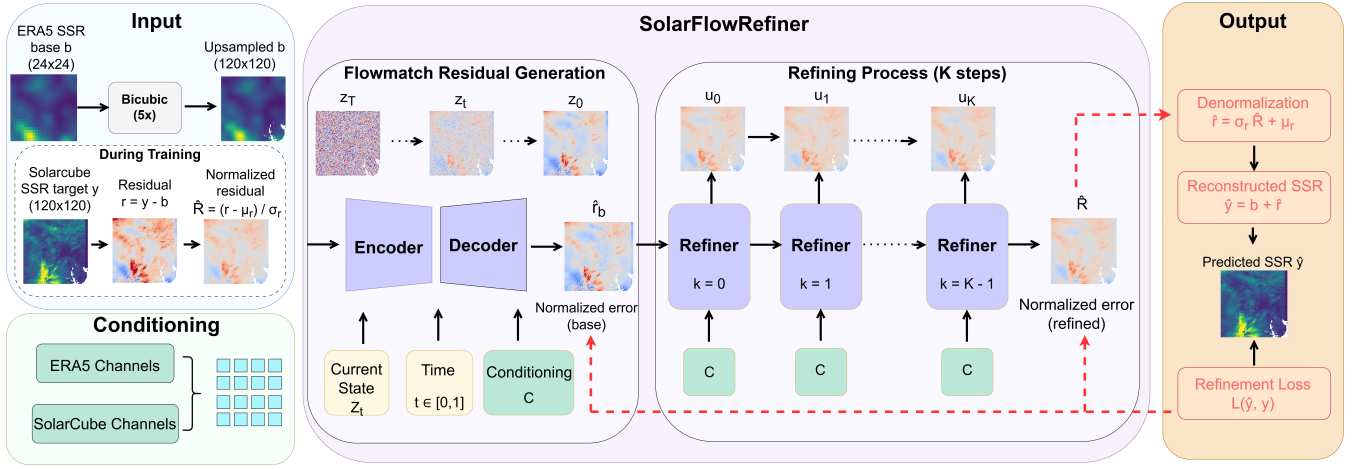}
\caption{Overview of SolarFlowRefiner. ERA5 radiative variables and SolarCube
auxiliary channels condition a FlowMatch residual generator, which predicts an
initial normalized residual relative to the upsampled ERA5 SSR baseline. A
PDE-style refiner then performs iterative correction updates over $K=8$
refinement levels before reconstructing the high-resolution SSR field. In
SolarFlowRefiner, the refinement loss is backpropagated through both the refiner
and the FlowMatch sampler, making generation refinement-aware.}
\label{fig:architecture}
\end{figure*}

\section{Problem Setup}
\label{sec:problem}
We study supervised downscaling from coarse ERA5 radiative fields to
high-resolution SolarCube SSR. Each sample contains conditioning inputs
$x \in \mathbb{R}^{C \times 120 \times 120}$ and a target SSR field
$y \in \mathbb{R}^{1 \times 120 \times 120}$. ERA5 variables are upsampled to the
SolarCube grid. Let $b$ denote the upsampled ERA5 SSR baseline. Rather than
predicting $y$ directly, we define the residual
\begin{equation}
    r = y - b .
\end{equation}
We normalize this residual using training-set statistics,
\begin{equation}
    \tilde r = \frac{r - \mu_r}{\sigma_r},
\end{equation}
and reconstruct physical SSR from a predicted normalized residual
$\hat{\tilde r}$ as
\begin{equation}
    \hat y = b + \sigma_r \hat{\tilde r} + \mu_r .
\end{equation}
This formulation preserves the coarse physical estimate from ERA5 and focuses
learning on high-resolution cloud-driven corrections.

The conditioning variables include five ERA5 radiative channels:
\texttt{era\_ssrd}, \texttt{era\_ssr}, \texttt{era\_ssrdc},
\texttt{era\_fdir}, and \texttt{era\_cdir}. SolarCube auxiliary channels include
\texttt{vis047}, \texttt{vis086}, \texttt{ir133}, \texttt{sza}, and
\texttt{cm}. Together, these channels provide both coarse radiative context and
high-resolution cloud information. For input-channel ablations, the ERA5 SSR
baseline $b$ remains the residual reference in all settings, while the ablation
changes only which channels are provided to the learned generator and refiner.

We use the split 70-10-20 among train-val-test respectively. Tiles
1--10, 12, and 14 pass the target-valid quality-control threshold and are used
for training and evaluation. Tiles 11, 13, and 15--19 are excluded because their
target-valid fraction is below 50\%. The split contains 32,159 training samples,
4,519 validation samples, and 8,994 test samples. Samples are assigned by whole
UTC-day blocks within each tile/month, so all hours from the same tile-month-day
remain in the same partition. This reduces leakage from near-duplicate cloud
scenes across train, validation, and test sets.

We report MAE and RMSE in physical irradiance units, SSIM \cite{ssim} for structural agreement, and LPIPS \cite{lpips} and FID \cite{fid} for perceptual and distributional fidelity.

\section{Method}
\label{sec:method}

SolarFlowRefiner is a learned predictor--corrector framework for
high-resolution surface solar radiation (SSR) reconstruction. The predictor is
a conditional FlowMatch model that generates an initial estimate of the
normalized SSR residual. The corrector is a multilevel denoising refiner that
progressively removes the structured errors remaining in that estimate. The
central design choice is to optimize these two components jointly: the
refinement objective is differentiated through the FlowMatch sampler, allowing
the generator to adapt to the downstream correction process rather than being
trained only as an independent predictor.

Let $x$ denote the conditioning variables, $\tilde r$ the normalized target
residual, and $b$ the upsampled ERA5 SSR baseline defined in
Section~\ref{sec:problem}. SolarFlowRefiner first generates a base residual
estimate $\tilde r_b$ and then applies $K$ refinement updates to obtain the final
normalized residual $\hat{\tilde r}$. The physical SSR prediction is reconstructed
as
\begin{equation}
    \hat y
    =
    b + \sigma_r \hat{\tilde r} + \mu_r .
\end{equation}

\subsection{Conditional FlowMatch Predictor}
\label{sec:flowmatch}

The base predictor models the conditional distribution of normalized SSR
residuals using rectified flow matching. Let
$z_0 \sim \mathcal{N}(0,I)$ denote a Gaussian source sample and let
$z_1=\tilde r$ denote the target residual. For a continuous flow time
$t\sim\mathcal{U}(0,1)$, we define the linear conditional probability path
\begin{equation}
    z_t=(1-t)z_0+t z_1 .
    \label{eq:fm_path}
\end{equation}
Along this path, the target transport velocity is constant:
\begin{equation}
    v^\star(z_t,z_0,z_1)=z_1-z_0.
\end{equation}
A conditional velocity network $f_\theta$ receives the current state $z_t$,
the conditioning variables $x$, and the flow time $t$. It is trained using
\begin{equation}
    \mathcal{L}_{\mathrm{FM}}(\theta)
    =
    \mathbb{E}_{x,\tilde r,z_0,t}
    \left[
        \left\|
            f_\theta(z_t,x,t)-(\tilde r-z_0)
        \right\|_1
    \right].
    \label{eq:fm_loss}
\end{equation}

At inference, generation follows the learned ordinary differential equation
\begin{equation}
    \frac{d z(t)}{dt}
    =
    f_\theta(z(t),x,t),
    \qquad
    z(0)=z_0.
    \label{eq:fm_ode}
\end{equation}
We numerically integrate Equation~\eqref{eq:fm_ode} from $t=0$ to $t=1$.
Using $M$ integration steps with time points
$0=t_0<\cdots<t_M=1$, an Euler update takes the form
\begin{equation}
    z^{(m+1)}
    =
    z^{(m)}
    +
    (t_{m+1}-t_m)
    f_\theta(z^{(m)},x,t_m).
    \label{eq:fm_sampling}
\end{equation}
The resulting base residual is
\begin{equation}
    \tilde r_b
    =
    S_\theta(z_0,x)
    =
    z^{(M)},
    \label{eq:base_residual}
\end{equation}
where $S_\theta$ denotes the differentiable FlowMatch sampling procedure.

The base prediction captures the overall residual field, but a finite-capacity
generator and few-step sampler may leave localized errors around cloud
boundaries, shadow transitions, and high-gradient irradiance regions. The
refinement stage is designed to correct these candidate-specific residual
errors without regenerating the entire field from noise.

\subsection{Prediction-Conditioned Refinement Path}
\label{sec:refinement_path}

A standard denoising model is commonly trained by perturbing the ground-truth
target directly. Such perturbations do not necessarily resemble the structured
errors produced by the FlowMatch generator. We instead construct the refiner's
training states from the current generated residual $\tilde r_b$. This makes the
training distribution explicitly dependent on the prediction that will be
refined at inference.

For refinement level $k\in\{0,\ldots,K-1\}$, define
\begin{equation}
    \alpha_k=\frac{k}{K-1}.
    \label{eq:alpha_schedule}
\end{equation}
We construct a prediction-conditioned state along the path from the current
FlowMatch residual to the target:
\begin{equation}
    c_k
    =
    (1-\alpha_k)\tilde r_b
    +
    \alpha_k\tilde r .
    \label{eq:refinement_path}
\end{equation}
The path begins at the generator prediction,
$c_0=\tilde r_b$, and terminates at the target,
$c_{K-1}=\tilde r$. Writing the current generator error as
\begin{equation}
    e_\theta=\tilde r_b-\tilde r,
\end{equation}
Equation~\eqref{eq:refinement_path} gives
\begin{equation}
    c_k-\tilde r
    =
    (1-\alpha_k)e_\theta.
    \label{eq:error_scaling}
\end{equation}
Thus, the refinement levels expose the corrector to progressively smaller
versions of the structured errors produced by the current generator, rather
than to arbitrary interpolation errors.

To improve robustness and train the corrector at multiple uncertainty levels,
we perturb each intermediate state using Gaussian noise:
\begin{equation}
    \bar c_k
    =
    c_k+\sigma_k\epsilon,
    \qquad
    \epsilon\sim\mathcal{N}(0,I).
    \label{eq:noisy_refinement_state}
\end{equation}
We use an exponentially decreasing noise schedule
\begin{equation}
    \sigma_k
    =
    \sigma_{\max}
    \left(
        \frac{\sigma_{\min}}{\sigma_{\max}}
    \right)^{\frac{k}{K-1}},
    \label{eq:noise_schedule}
\end{equation}
so early refinement levels cover larger perturbations around the base
prediction, while later levels focus on small corrections near the target.

Combining Equations~\eqref{eq:error_scaling} and
\eqref{eq:noisy_refinement_state}, the refiner observes
\begin{equation}
    \bar c_k-\tilde r
    =
    (1-\alpha_k)e_\theta+\sigma_k\epsilon.
    \label{eq:structured_plus_noise}
\end{equation}
Its input therefore contains both the structured error of the current generator
and a controlled stochastic perturbation.

\subsection{Multilevel Denoising Refiner}
\label{sec:pde_refiner}

The PDE-style refiner $g_\phi$ receives the noisy intermediate state, the
original conditioning variables, the base FlowMatch prediction, and an
embedding of the refinement level:
\begin{equation}
    \hat r_k
    =
    g_\phi(\bar c_k,x,\tilde r_b,k).
    \label{eq:refiner_prediction}
\end{equation}
The base residual is supplied separately because it identifies the prediction
being corrected and allows the refiner to distinguish candidate-specific
errors from the Gaussian perturbation introduced during training.

The network predicts the clean target residual rather than the added noise.
For a sampled refinement level $k$, its loss is
\begin{align}
    \mathcal{L}_{\mathrm{ref}}^{(k)}
    =
    &\left\|\hat r_k-\tilde r\right\|_1
    +
    \lambda_{\mathrm{mse}}
    \left\|\hat r_k-\tilde r\right\|_2^2
    \nonumber\\
    &+
    \lambda_{\nabla}
    \left\|
        \nabla \hat r_k-\nabla\tilde r
    \right\|_1 .
    \label{eq:refinement_loss}
\end{align}
The first term provides robust pixel-level supervision, while the MSE term
places additional emphasis on large residual errors. The gradient-consistency
term encourages the reconstruction to preserve spatial transitions associated
with cloud boundaries and localized irradiance changes. The complete refinement
objective averages over training samples, FlowMatch source noise, refinement
levels, and perturbation noise:
\begin{equation}
    \mathcal{L}_{\mathrm{ref}}(\theta,\phi)
    =
    \mathbb{E}_{x,\tilde r,z_0,k,\epsilon}
    \left[
        \mathcal{L}_{\mathrm{ref}}^{(k)}
    \right].
    \label{eq:expected_refinement_loss}
\end{equation}
The dependence on $\theta$ arises because
$\tilde r_b=S_\theta(z_0,x)$ determines both the refinement path and one of the
refiner inputs.

\subsection{Iterative Refinement at Inference}
\label{sec:refinement_inference}

At inference, the refinement process begins from the generated residual
\begin{equation}
    u_0=\tilde r_b.
\end{equation}
For $k=0,\ldots,K-1$, the refiner predicts a clean residual estimate
\begin{equation}
    \hat r_k
    =
    g_\phi(u_k,x,\tilde r_b,k),
    \label{eq:inference_prediction}
\end{equation}
and the current state is moved toward this prediction:
\begin{equation}
    u_{k+1}
    =
    u_k+\eta_k(\hat r_k-u_k).
    \label{eq:refinement_update}
\end{equation}
Here, $\eta_k\in(0,1]$ controls the correction strength. We use a constant
$\eta_k=\eta$ in our experiments. Unlike training, no ground-truth residual is
available and no prediction-to-target interpolation is constructed at
inference. The refiner instead follows the sequence of level-conditioned
correction operators learned from Equation~\eqref{eq:refinement_path}. The final
normalized residual is
\begin{equation}
    \hat{\tilde r}=u_K.
    \label{eq:final_residual}
\end{equation}

The distinction between Equations~\eqref{eq:refinement_path} and
\eqref{eq:refinement_update} is important. The former defines the supervised
training distribution using the known target, whereas the latter defines the
test-time correction trajectory using only the current estimate and available
conditioning information.

\subsection{Refinement-Aware Joint Optimization}
\label{sec:joint_training}

A frozen generator--refiner cascade optimizes its two stages independently. The
generator is first trained with Equation~\eqref{eq:fm_loss}, after which its
output is treated as a fixed input when training the refiner. In that setting,
the refinement objective cannot influence the states produced by the generator.

SolarFlowRefiner instead runs the differentiable sampler
$S_\theta$ inside the refinement training loop. Its coupled objective is
\begin{equation}
    \mathcal{L}_{\mathrm{joint}}(\theta,\phi)
    =
    \lambda_{\mathrm{FM}}\mathcal{L}_{\mathrm{FM}}(\theta)
    +
    \lambda_{\mathrm{ref}}
    \mathcal{L}_{\mathrm{ref}}(\theta,\phi).
    \label{eq:joint_objective}
\end{equation}
Because the base residual appears in both the noisy path state and the refiner
conditioning, the refinement gradient with respect to the generator can be
written schematically as
\begin{align}
    \nabla_\theta\mathcal{L}_{\mathrm{ref}}
    =
    \Bigg[
        &(1-\alpha_k)
        \frac{\partial\mathcal{L}_{\mathrm{ref}}}
             {\partial\bar c_k}
        +
        \frac{\partial\mathcal{L}_{\mathrm{ref}}}
             {\partial\tilde r_b}
    \Bigg]
    \frac{\partial S_\theta(z_0,x)}
         {\partial\theta}.
    \label{eq:coupled_gradient}
\end{align}
The first term propagates through the prediction-conditioned refinement state,
while the second propagates through the explicit base-residual conditioning of
the refiner. Therefore,
\begin{equation}
    \nabla_\theta\mathcal{L}_{\mathrm{ref}}\neq 0
\end{equation}
in SolarFlowRefiner.

This coupling does not encourage the generator to produce a less accurate
prediction. Rather, it augments the standalone FlowMatch objective with
information about how the generated state behaves under downstream correction.
The generator and refiner consequently co-adapt: the generator supplies the
states defining the refiner's training distribution, while the refinement
objective discourages generator outputs that lead to large final correction
errors.

\subsection{Model Variants}
\label{sec:model_variants}

We evaluate the following four flow-based configurations.

\paragraph{FlowMatch.}
This is the standalone conditional residual generator described in
Section~\ref{sec:flowmatch}. Its output $\tilde r_b$ is used directly for SSR
reconstruction, without downstream correction.

\paragraph{FlowRefiner-ODE.}
This post-hoc variant freezes the pretrained FlowMatch generator and replaces
the denoising corrector with an ODE-style velocity model. The velocity model
learns a deterministic correction trajectory from the fixed FlowMatch residual
toward the target residual. At inference, this correction field is numerically
integrated from $\tilde r_b$ to obtain the final prediction. Gradients from the
ODE refiner are not propagated into the FlowMatch generator.

\paragraph{FlowRefiner-PDE.}
This variant uses the multilevel denoising refiner defined in
Sections~\ref{sec:refinement_path}--\ref{sec:refinement_inference}, but trains it
on residuals produced by a frozen FlowMatch generator. The base predictions may
be precomputed and cached. Formally, the refinement path is constructed using
\begin{equation}
    \tilde r_b
    =
    \operatorname{sg}
    \left[
        S_\theta(z_0,x)
    \right],
\end{equation}
where $\operatorname{sg}[\cdot]$ denotes the stop-gradient operator. Hence,
\begin{equation}
    \nabla_\theta\mathcal{L}_{\mathrm{ref}}=0.
\end{equation}

\paragraph{SolarFlowRefiner.}
This is the proposed refinement-aware model. The FlowMatch prediction is
generated online, the prediction-conditioned refinement path is constructed
from the current generator output, and the refinement loss is differentiated
through the sampler. SolarFlowRefiner therefore differs from
FlowRefiner-PDE in training coupling rather than in the inference-time
refinement architecture.

\begin{table*}[!t]
\centering
\caption{Test-set SSR downscaling performance. Lower is better for MAE, RMSE,
LPIPS, and FID; higher is better for SSIM. All metrics are computed after
reconstructing the physical SSR field from the predicted normalized residual.}
\label{tab:main_results}
\normalsize
\renewcommand{\arraystretch}{1.25}
\begin{tabular*}{\textwidth}{@{\extracolsep{\fill}}lccccc@{}}
\toprule
Method & MAE$\downarrow$ & RMSE$\downarrow$ & SSIM$\uparrow$ & LPIPS$\downarrow$ & FID$\downarrow$ \\
\midrule
Bicubic           & 137.22 & 169.61 & 0.339 & 0.664 & 290.15 \\
EDSR              & 29.53  & 42.71  & 0.675 & 0.187 & 46.03 \\
RCAN              & 14.81  & 24.11  & 0.724 & 0.177 & 44.92 \\
SwinIR            & 16.71  & 26.91  & 0.712 & 0.183 & 39.21 \\
HAT               & 15.25  & 24.07  & 0.726 & 0.171 & 37.73 \\
ResShift          & 13.94  & 20.68  & 0.725 & 0.159 & 36.47 \\
FlowMatch         & 14.88  & 23.30  & 0.742 & 0.154 & \textbf{35.42} \\
FlowRefiner-ODE   & 14.60  & 22.85  & 0.743 & 0.156 & 36.43 \\
FlowRefiner-PDE   & 13.58  & 20.31  & 0.727 & 0.161 & 37.49 \\
SolarFlowRefiner  & \textbf{11.13} & \textbf{17.11} & \textbf{0.827} & \textbf{0.137} & 36.23 \\
\bottomrule
\end{tabular*}
\end{table*}

\begin{table}[!t]
\centering
\caption{SolarFlowRefiner input-channel ablation. The ERA5 SSR baseline is used
for residual reconstruction in all settings; the ablation changes only the
conditioning channels provided to the learned generator and refiner.}
\label{tab:channel_ablation}
\small
\setlength{\tabcolsep}{3.2pt}
\renewcommand{\arraystretch}{1.08}
\begin{tabular}{lccccc}
\toprule
Conditioning & MAE$\downarrow$ & RMSE$\downarrow$ & SSIM$\uparrow$ & LPIPS$\downarrow$ & FID$\downarrow$ \\
\midrule
ERA5 only        & 21.82 & 36.21 & 0.651 & 0.172 & 41.90 \\
SolarCube only   & 17.31 & 24.70 & 0.710 & 0.155 & 37.10 \\
ERA5 + SolarCube & \textbf{11.13} & \textbf{17.11} & \textbf{0.827} & \textbf{0.137} & \textbf{36.23} \\
\bottomrule
\end{tabular}
\end{table}

\begin{figure*}[t]
\centering
\includegraphics[width=\textwidth]{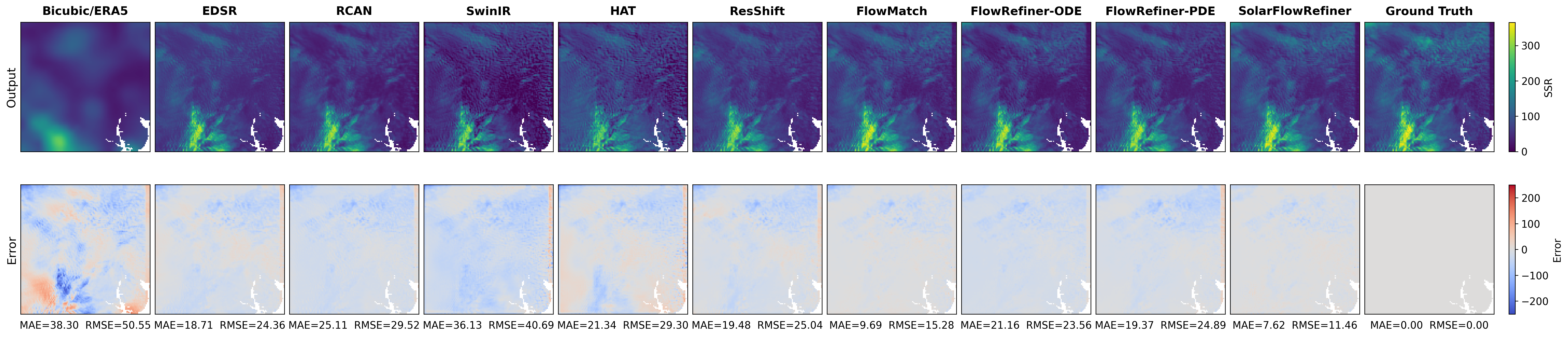}
\caption{Qualitative SSR reconstruction and error maps for a held-out test
sample. The top row shows reconstructed SSR fields for Bicubic/ERA5,
deterministic baselines, generative baselines, refinement variants, and the
SolarCube ground truth. The bottom row shows the corresponding error maps
computed as prediction minus ground truth, with per-sample MAE and RMSE reported
below each panel.}
\label{fig:qual_ssr}
\end{figure*}

\begin{figure*}[t]
\centering
\includegraphics[width=\textwidth]{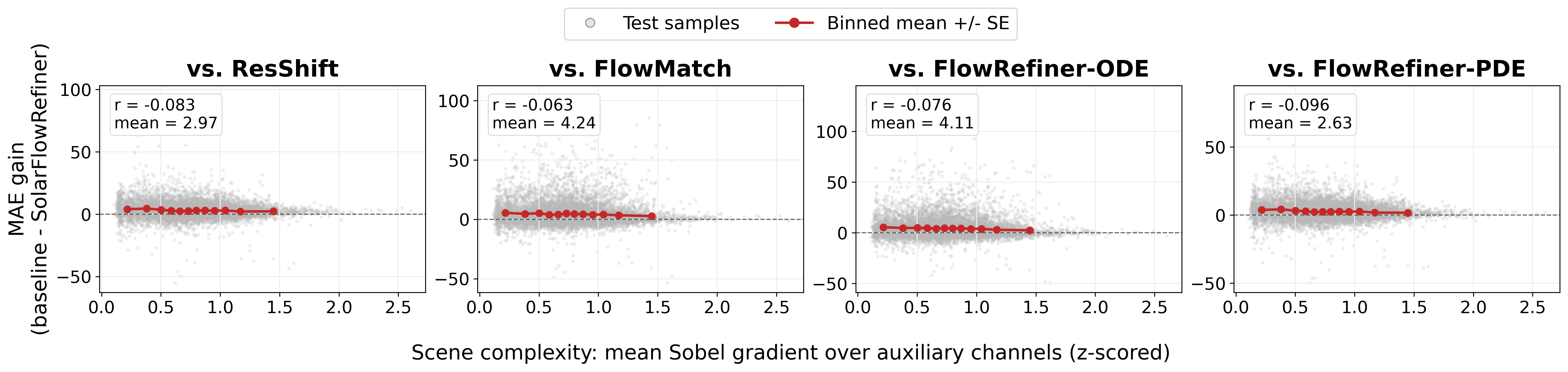}
\caption{Performance gain versus scene complexity. Scene complexity is computed
as the mean Sobel gradient magnitude over z-scored SolarCube auxiliary channels
\texttt{vis047}, \texttt{vis086}, \texttt{ir133}, and \texttt{cm}. Each gray
point is a held-out test sample. Red curves show binned mean MAE gain with
standard-error bars, where gain is baseline MAE minus SolarFlowRefiner MAE.
Positive values indicate that SolarFlowRefiner has lower error than the
baseline.}
\label{fig:gain_complexity}
\end{figure*}

\section{Experimental Setup}
\label{sec:experiments}

We conduct extensive experiments to address the following research questions:

\begin{itemize}
    \item \textbf{RQ1: Baseline performance.}
    How effectively do deterministic, residual-shift, and flow-based methods
    reconstruct high-resolution SSR residuals?

    \item \textbf{RQ2: Post-hoc refinement.}
    Does iterative refinement improve the standalone FlowMatch prediction, and
    how do ODE-style and PDE-style refinement dynamics compare?

    \item \textbf{RQ3: Refinement-aware coupling.}
    Does end-to-end optimization of the generator and refiner improve over
    independently trained post-hoc refinement?

    \item \textbf{RQ4: Conditioning information.}
    How do ERA5 radiative variables and SolarCube satellite channels contribute
    to SolarFlowRefiner's downscaling performance?
\end{itemize}

\textbf{Dataset and evaluation.}
We evaluate on the ERA5--SolarCube split described in the problem setup. All
models are trained on the 32,159-sample training set, selected using validation
performance, and evaluated on the 8,994-sample held-out test set. Metrics are
computed after reconstructing physical SSR from the predicted normalized
residual.

\textbf{Baselines.}
We compare against Bicubic interpolation of ERA5 SSR, deterministic residual
super-resolution baselines EDSR and RCAN, attention-based SwinIR and
HAT baselines, ResShift, standalone FlowMatch, FlowRefiner-ODE,
FlowRefiner-PDE, and SolarFlowRefiner. EDSR, RCAN, SwinIR, and HAT are included
as representative deterministic or attention-based super-resolution models
\citep{edsr,rcan,swinir,hat}. ResShift represents residual-shift generative
super-resolution \citep{resshift}. FlowMatch, FlowRefiner-ODE, and
FlowRefiner-PDE isolate the effects of generation and post-hoc refinement, while
SolarFlowRefiner tests the value of refinement-aware end-to-end coupling.

\textbf{Implementation details.}
The generative models use a multiscale residual U-Net backbone. During end-to-end training, FlowMatch sampling uses an 8-step Euler solver for memory and runtime efficiency. The refiner learning rate is $2\times10^{-5}$ and the FlowMatch learning rate is$2\times10^{-6}$. The refiner uses $K=8$ refinement levels with a noise schedule decreasing from 0.35 to 0.01. We use AdamW with batch size 4 and train for 100 epochs. In the refinement loss, we set $\lambda_{\mathrm{mse}}=0.1$ and $\lambda_{\nabla}=0.05$, and use refinement strength $\eta=1.0$. All training is performed on Nvidia RTX 5060 GPU.

\textbf{Input-channel ablation.}
To answer RQ4, we evaluate SolarFlowRefiner under three conditioning settings:
ERA5-only, SolarCube-only, and ERA5+SolarCube. The ERA5-only setting includes
\texttt{era\_ssrd}, \texttt{era\_ssr}, \texttt{era\_ssrdc},
\texttt{era\_fdir}, and \texttt{era\_cdir}. The SolarCube-only setting includes
\texttt{vis047}, \texttt{vis086}, \texttt{ir133}, \texttt{sza}, and
\texttt{cm}. The combined setting uses all ten channels. In every setting, the
upsampled ERA5 SSR field remains the residual baseline $b$ used for
reconstruction.

\section{Results}

\textbf{Main quantitative comparison.}
Table~\ref{tab:main_results} reports the held-out test-set comparison. Bicubic
upsampling of ERA5 SSR has high error because the coarse reanalysis field cannot
resolve localized cloud-shadow and cloud-edge structure. Deterministic
super-resolution models reduce this error substantially, but their gains
saturate once the model must recover sharper, high-frequency SSR residuals. The
generative and refinement-based models provide a stronger reconstruction of
these residual fields.

SolarFlowRefiner achieves the best MAE, RMSE, SSIM, and LPIPS among the methods
in Table~\ref{tab:main_results}. Relative to the strongest PDE-refinement
baseline, FlowRefiner-PDE, SolarFlowRefiner reduces MAE from 13.58 to 11.13
and RMSE from 20.31 to 17.11, corresponding to approximately 18\% lower MAE and
16\% lower RMSE. It also improves SSIM from 0.727 to 0.827, indicating better
structural agreement with SolarCube. Compared with standalone FlowMatch,
SolarFlowRefiner reduces MAE by about 25\% and RMSE by about 27\%, showing that
the improvement is not due only to using a flow-based generator, but to coupling
generation with downstream refinement. FID is comparable to the strongest
generative baselines, while LPIPS is lowest for SolarFlowRefiner, suggesting
that the proposed model improves local perceptual structure without degrading
distribution-level realism.

\textbf{Input-channel ablation.}
Table~\ref{tab:channel_ablation} reports the SolarFlowRefiner input-channel
ablation. We include this ablation only for the proposed model so
that the analysis isolates the role of conditioning sources without conflating
input selection with architecture changes.

The ablation confirms that both sources of conditioning information are useful.
Using ERA5 channels alone gives substantially higher error, because the model
receives physically meaningful radiative variables but little high-resolution
cloud structure. SolarCube-only conditioning improves over ERA5-only
conditioning, reducing MAE from 21.82 to 17.31, which indicates that the
high-resolution auxiliary channels carry important cloud-boundary information.
The full ERA5+SolarCube setting performs best across all metrics, reducing MAE
by about 49\% relative to ERA5-only conditioning and about 36\% relative to
SolarCube-only conditioning. This supports the residual formulation: ERA5
provides the coarse physical baseline, while SolarCube auxiliary channels guide
the fine-scale correction.

\textbf{Qualitative reconstruction.}
Figure~\ref{fig:qual_ssr} shows a representative held-out sample. Bicubic/ERA5
preserves only broad radiative structure and misses the fine cloud-driven
texture visible in the ground truth. Deterministic baselines recover more local
detail but still leave structured errors near high-gradient cloud boundaries.
FlowMatch and the post-hoc refinement variants reduce these errors, but
SolarFlowRefiner produces the lowest per-sample MAE and RMSE in this example.
The error maps show that refinement-aware training reduces both broad residual
bias and localized cloud-edge mistakes relative to the post-hoc refiners.

\textbf{Scene complexity analysis.}
Figure~\ref{fig:gain_complexity} evaluates whether SolarFlowRefiner's gains are
concentrated in scenes with stronger spatial structure. For each test sample, we
measure scene complexity as the mean Sobel gradient magnitude over the
SolarCube auxiliary channels \texttt{vis047}, \texttt{vis086}, \texttt{ir133},
and \texttt{cm}. We then compute gain as the baseline MAE minus the
SolarFlowRefiner MAE, so positive values indicate that SolarFlowRefiner is
better.

SolarFlowRefiner has positive average gain against all four baselines:
2.97 MAE over ResShift, 4.24 over FlowMatch, 4.11 over FlowRefiner-ODE, and
2.63 over FlowRefiner-PDE. The Pearson correlations between complexity and gain
are weakly negative, ranging from $r=-0.063$ to $r=-0.096$. This indicates that
the improvement is broad rather than restricted to a narrow complexity regime.
The binned means remain above zero across most of the complexity range, including
against FlowRefiner-PDE, supporting the claim that refinement-aware coupling
provides a consistent advantage over post-hoc refinement.

\section{Conclusion and Future Work}
In this paper, we introduced SolarFlowRefiner for multisource surface solar radiation downscaling. The framework addresses the stage-wise mismatch between generation and post-hoc refinement by constructing prediction-conditioned refinement states from the current generator output and propagating the refinement objective through the differentiable sampler. This aligns generation and correction with the final SSR reconstruction objective.
Experiments on the day-blocked ERA5--SolarCube benchmark show that SolarFlowRefiner consistently outperforms standalone generation and frozen refinement variants. Relative to the strongest post-hoc PDE-style refiner, it reduces MAE and RMSE by approximately 18\% and 16\%, respectively, while improving structural and perceptual fidelity. The conditioning ablation further confirms that coarse ERA5 radiative information and high-resolution SolarCube observations provide complementary information for fine-scale reconstruction.
These findings point to a broader class of geophysical inverse problems in which coarse, physically consistent products capture the large-scale system state, while remote-sensing observations reveal localized spatial heterogeneity. Future work will investigate refinement-aware generation as a transferable predictor--corrector principle for such problems, with generation recovering a plausible global field and refinement resolving observation-informed fine-scale structure. We will first extend this framework to land-surface temperature, precipitation, and soil-moisture reconstruction, and then examine cross-region and cross-sensor transfer toward more general AI methods for multisource Earth observation and environmental decision-making.

\bibliography{aaai2027}

\appendix

\section{Dataset Construction Details}
\label{app:dataset}

Our benchmark is constructed for hourly ERA5-to-SolarCube surface solar
radiation (SSR) downscaling. Each sample contains a coarse reanalysis-derived
radiative state from ERA5 and a high-resolution SolarCube target on a
$120\times120$ grid. ERA5 variables are spatially upsampled to the SolarCube
grid before being passed to the model. The high-resolution target is the hourly
mean SolarCube SSR field, computed from the available 15-minute SolarCube frames
within the hour. We exclude nighttime samples using a minimum ERA5 SSRD mean
threshold of 10.0, and we discard samples with target-valid fraction below the
quality-control threshold used in the main split.

The conditioning tensor contains ten channels. The five ERA5 radiative channels
are \texttt{era\_ssrd}, \texttt{era\_ssr}, \texttt{era\_ssrdc},
\texttt{era\_fdir}, and \texttt{era\_cdir}. The five SolarCube auxiliary
channels are \texttt{vis047}, \texttt{vis086}, \texttt{ir133}, \texttt{sza},
and \texttt{cm}. The prediction target is the SolarCube SSR field
\texttt{solarcube\_ssr\_hourly}. All learned models predict the normalized SSR
residual rather than the HR field directly. The residual is the SolarCube target
minus the upsampled ERA5 SSR baseline, standardized using training-set residual
statistics. Physical SSR is reconstructed by adding the predicted residual
correction back to the ERA5 baseline. This residual formulation preserves the
coarse physical estimate from ERA5 and focuses learning on high-resolution
cloud-driven corrections.

We use tiles 1--10, 12, and 14, which pass the target-valid quality-control
threshold. Tiles 11, 13, and 15--19 are excluded because their target-valid
fraction is below 50\%. Samples are split by whole UTC-day blocks within each
tile/month, so all hours from the same tile-month-day remain in the same
partition. This reduces leakage from near-duplicate cloud scenes across the
training, validation, and test sets. The final split uses a 70/10/20
train/validation/test ratio and is summarized in Table~\ref{tab:app_split}.

\begin{table}[h]
\centering
\caption{Dataset split used for model selection and final evaluation.}
\label{tab:app_split}
\small
\begin{tabular}{lcc}
\toprule
Split & Samples & Fraction \\
\midrule
Train & 32,159 & 70\% \\
Validation & 4,519 & 10\% \\
Test & 8,994 & 20\% \\
\bottomrule
\end{tabular}
\end{table}

\section{Site-Wise Spatial Performance Analysis}
\label{app:spatial}

We further analyze whether the relative benefit of SolarFlowRefiner is uniform
across the spatial sites. This analysis uses the same 8,994 held-out
test samples and the same per-sample MAE values used for the main quantitative
comparison. We focus on spatial variation rather than UTC-hour variation because
the sites span multiple longitudes; the same UTC hour can correspond to different
local solar times at different stations. This makes a raw UTC morning/afternoon
analysis difficult to interpret without additional solar-time normalization.

\begin{figure*}[!t]
\centering
\includegraphics[width=0.95\textwidth]{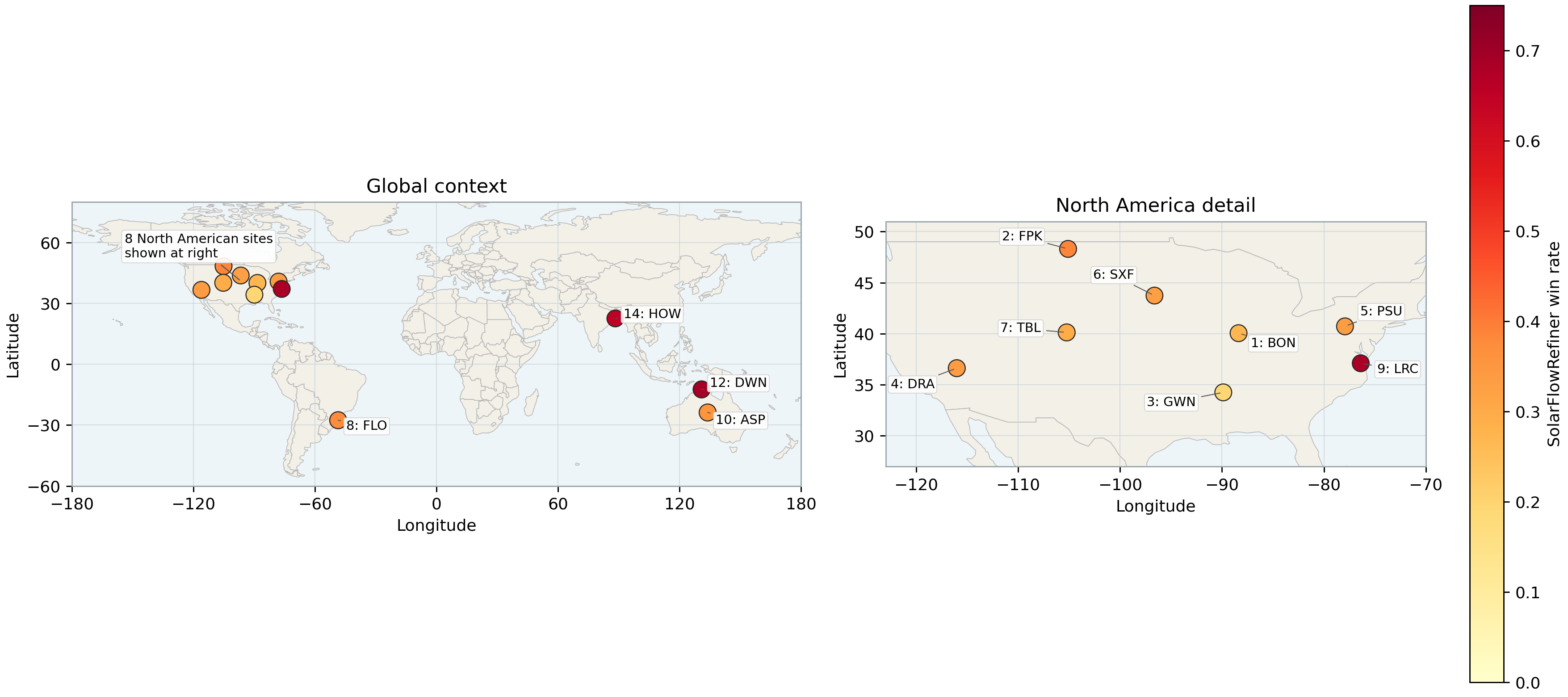}
\caption{Global distribution of the 12 test tiles/sites. Marker color
shows the fraction of test samples at each site for which SolarFlowRefiner has
the lowest MAE among the learned comparison models. The right panel expands the
North American cluster.}
\label{fig:app_station_map}
\end{figure*}

Figure~\ref{fig:app_station_map} shows the spatial distribution of the test
sites. Marker color indicates the per-sample MAE win rate of SolarFlowRefiner at
each site, where the competing methods are ResShift, FlowMatch,
FlowRefiner-ODE, and FlowRefiner-PDE. The eight North American sites are
shown separately in an inset because they are geographically clustered. Station
codes, station names, coordinates, and test sample counts are listed in
Table~\ref{tab:app_station_metadata}. The map uses Natural Earth country
boundaries for geographic context; station labels and coordinates follow the
BSRN/PANGAEA station metadata conventions.

\begin{table*}[!t]
\centering
\caption{Test sites used in the spatial analysis.}
\label{tab:app_station_metadata}
\small
\resizebox{\textwidth}{!}{%
\begin{tabular}{rlllrrr}
\toprule
Tile & Code & Station & Region & Latitude & Longitude & Test samples \\
\midrule
1 & BON & Bondville & Illinois, USA & 40.052 & -88.373 & 745 \\
2 & FPK & Fort Peck & Montana, USA & 48.308 & -105.102 & 746 \\
3 & GWN & Goodwin Creek & Mississippi, USA & 34.255 & -89.873 & 744 \\
4 & DRA & Desert Rock & Nevada, USA & 36.624 & -116.019 & 756 \\
5 & PSU & Penn State University & Pennsylvania, USA & 40.720 & -77.931 & 743 \\
6 & SXF & Sioux Falls & South Dakota, USA & 43.734 & -96.623 & 736 \\
7 & TBL & Table Mountain & Colorado, USA & 40.126 & -105.238 & 756 \\
8 & FLO & Florianopolis & Brazil & -27.605 & -48.523 & 764 \\
9 & LRC & Langley Research Center & Virginia, USA & 37.104 & -76.387 & 743 \\
10 & ASP & Alice Springs & Northern Territory, Australia & -23.798 & 133.888 & 763 \\
12 & DWN & Darwin Met Office & Northern Territory, Australia & -12.424 & 130.893 & 765 \\
14 & HOW & Howrah & India & 22.554 & 88.306 & 733 \\
\bottomrule
\end{tabular}%
}
\end{table*}

Table~\ref{tab:app_spatial_mae} reports site-wise mean MAE. SolarFlowRefiner
has the lowest mean MAE at every site, indicating that the coupled
FlowMatch--refiner model improves the average reconstruction error across the
spatial domain rather than only at a small subset of stations. The size of the
benefit, however, is spatially heterogeneous. The largest absolute reductions
occur at LRC, DWN, and HOW, where the competing methods have substantially
higher mean errors and SolarFlowRefiner also wins most individual samples.

\begin{table*}[!t]
\centering
\caption{Per-site mean MAE in $\mathrm{W\,m^{-2}}$ for the
learned models. Lower is better.}
\label{tab:app_spatial_mae}
\small
\resizebox{\textwidth}{!}{%
\begin{tabular}{rlrrrrr}
\toprule
Tile & Site & ResShift & FlowMatch & FlowRefiner-ODE & FlowRefiner-PDE & SolarFlowRefiner \\
\midrule
1 & BON & 11.87 & 12.03 & 11.62 & 11.31 & \textbf{10.03} \\
2 & FPK & 17.31 & 18.72 & 19.07 & 16.45 & \textbf{13.55} \\
3 & GWN & 10.78 & 10.22 & 10.16 & 10.28 & \textbf{9.19} \\
4 & DRA & 12.44 & 12.24 & 12.56 & 12.69 & \textbf{10.95} \\
5 & PSU & 12.53 & 11.78 & 11.59 & 12.43 & \textbf{9.66} \\
6 & SXF & 17.36 & 17.56 & 17.39 & 16.66 & \textbf{14.35} \\
7 & TBL & 14.51 & 13.74 & 13.76 & 13.88 & \textbf{12.09} \\
8 & FLO & 11.98 & 11.32 & 11.03 & 11.82 & \textbf{9.80} \\
9 & LRC & 15.73 & 19.00 & 17.56 & 15.78 & \textbf{11.54} \\
10 & ASP & 13.04 & 12.58 & 12.21 & 12.29 & \textbf{10.26} \\
12 & DWN & 13.76 & 16.22 & 16.16 & 13.77 & \textbf{9.91} \\
14 & HOW & 15.97 & 23.15 & 22.09 & 15.60 & \textbf{12.23} \\
\bottomrule
\end{tabular}%
}
\end{table*}

\begin{figure*}[!t]
\centering
\includegraphics[width=\textwidth,height=\textheight,keepaspectratio]{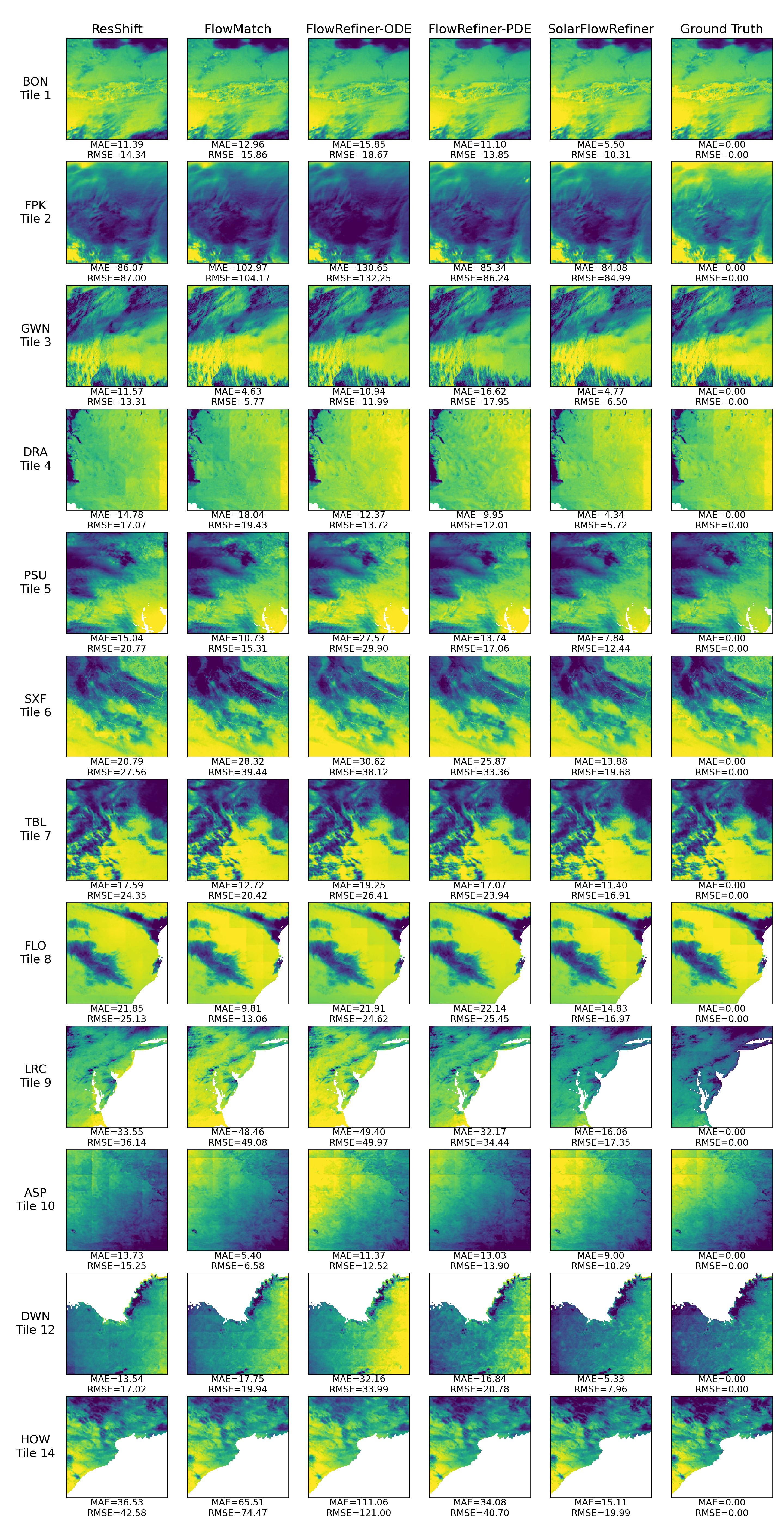}
\caption{Qualitative comparison across test sites. Each row shows one sample from a different tile/site, and columns compare ResShift,
FlowMatch, FlowRefiner-ODE, FlowRefiner-PDE, SolarFlowRefiner, and the
SolarCube ground truth. MAE and RMSE are reported below each reconstruction in
physical SSR units; the ground-truth column has zero error by definition.}
\label{fig:app_tile_qualitative_grid}
\end{figure*}

Figure~\ref{fig:app_tile_qualitative_grid} provides one qualitative example
from each site. The same models used in the spatial MAE analysis are
shown next to the SolarCube ground truth, with per-panel MAE and RMSE listed
below each reconstruction.

Overall, the spatial analysis supports two conclusions. First, the coupled
SolarFlowRefiner model improves mean reconstruction accuracy across all
sites in this split. Second, the strength of the improvement is site dependent:
the largest reductions occur at LRC, DWN, and HOW, while GWN and several central
U.S. stations show smaller gains. This spatial heterogeneity is consistent with
the interpretation that refinement is most helpful when the base generator
leaves structured cloud-related errors that can be corrected by a
prediction-conditioned refiner.

\section{Implementation and Training Details}
\label{app:implementation}

All learned models operate on $120\times120$ samples and predict normalized
residuals. Unless otherwise stated, models use all ten conditioning channels,
are selected using validation performance, and are evaluated on the
test set after reconstructing physical SSR. We use EMA checkpoints for the
generative and refinement models, with decay 0.9999, and use Adam or AdamW
optimization with gradient clipping at 1.0 where implemented.
All reported quantitative results use one final run for each model
configuration. The day-block split is generated with random seed 42, and model
training scripts use seed 42 for pseudorandom initialization and dataloader
control where applicable. Experiments were run in a Python/PyTorch CUDA
environment on an Nvidia RTX 5060 GPU. The anonymous code appendix includes the
preprocessing scripts, split manifests, model source code, run scripts, and a
\texttt{requirements.txt} file listing the required Python packages, including
PyTorch, NumPy, pandas, xarray, h5py, Pillow, and scikit-image.

\paragraph{Deterministic baselines.}
EDSR is trained as a residual CNN with 64 feature channels and 16 residual
blocks. RCAN is trained with 64 channels, 4 residual groups, and 4 residual
blocks per group. Both use batch size 16, learning rate $10^{-4}$, early
stopping on validation performance, and 100 training epochs. SwinIR and HAT are
included as attention-based super-resolution baselines. The SwinIR run uses
embedding dimension 60, window size 7, four stages with depths
$(6,6,6,6)$, heads $(6,6,6,6)$, batch size 8, learning rate
$2\times10^{-5}$, L1 loss, and gradient clipping at 1.0. The HAT run uses
embedding dimension 48, window size 6, depths $(2,2,2)$, heads $(4,4,4)$,
batch size 8, learning rate $3\times10^{-5}$, L1 loss, and gradient clipping at
1.0.

\paragraph{ResShift.}
The ResShift baseline uses a U-Net-style residual backbone with inner
channel 64, channel multipliers $(1,2,4,8,16)$, and two residual blocks per
stage. It is trained for 100 epochs with batch size 2, learning rate
$5\times10^{-5}$, weight decay $10^{-4}$, and EMA decay 0.9999. The residual
shift process uses $T=15$ transition steps with $\kappa=1.0$.

\paragraph{FlowMatch.}
The standalone FlowMatch generator uses a multiscale residual U-Net backbone
with base channel 64, channel multipliers $(1,2,4,8)$, and two residual blocks
per stage. It learns the rectified-flow velocity between Gaussian
source noise and the normalized SSR residual using an
$\ell_1$ velocity objective and uniform flow-time sampling,
$t \sim \mathcal{U}(0,1)$. The standalone FlowMatch baseline is trained
for 100 epochs with batch size 4, learning rate $2\times10^{-5}$, and EMA
decay 0.9999.

\paragraph{Post-hoc refiners.}
FlowRefiner-ODE keeps the pretrained FlowMatch base prediction fixed and trains
an ODE-style correction refiner. The refiner uses base channel 64, channel
multipliers $(1,2,4,8)$, two residual blocks per stage, 8 Heun refinement steps,
batch size 8, learning rate $2\times10^{-5}$, weight decay $10^{-4}$, and EMA
decay 0.9999. Its loss combines an L1 velocity term, endpoint supervision with
weight 0.5, and gradient consistency with weight 0.05.

FlowRefiner-PDE is the non-ODE post-hoc refiner built on the same pretrained
FlowMatch generator. It keeps the FlowMatch generator frozen, uses cached
FlowMatch base residual predictions as the initial state, and appends this base
prediction as an additional conditioning channel. Unlike FlowRefiner-ODE, which
learns a velocity field and integrates an ODE-style update, FlowRefiner-PDE uses
an iterative denoising-style correction. At each refinement level, the model
predicts the clean normalized residual from a state sampled along the
base-to-target path, and inference repeatedly moves the current residual toward
this predicted clean residual. The refiner uses base channel 64, channel
multipliers $(1,2,4,8)$, two residual blocks per stage, 8 refinement levels,
an exponentially decreasing noise schedule from $\sigma_{\max}=0.35$ to
$\sigma_{\min}=0.01$, refinement strength $\eta=1.0$, batch size 4, learning
rate $2\times10^{-5}$, weight decay $10^{-4}$, and EMA decay 0.9999. Its loss
combines L1 denoising, an MSE term with weight 0.1, and gradient consistency
with weight 0.05.

\paragraph{SolarFlowRefiner.}
SolarFlowRefiner is trained jointly from random initialization. The FlowMatch residual generator and PDE-style refiner are optimized together from the start. Each training batch applies the FlowMatch velocity objective, samples the current generator with an 8-step Euler solver, and trains the refiner on prediction-conditioned states
along the base-to-target path. The refinement loss combines L1 reconstruction,
an MSE term with weight 0.1, and gradient consistency with weight 0.05, and is
backpropagated through the differentiable FlowMatch sampler. The generator and
refiner both use base channel 64, channel multipliers $(1,2,4,8)$, and two
residual blocks per stage. The FlowMatch generator uses learning rate
$2\times10^{-6}$ and the refiner uses learning rate $2\times10^{-5}$ as
separate AdamW parameter groups. We use batch size 4, weight decay $10^{-4}$, EMA decay 0.9999, 100 training epochs, uniform FlowMatch time sampling
$t \sim \mathcal{U}(0,1)$, $K=8$ refinement levels, a noise
schedule from $\sigma_{\max}=0.35$ to $\sigma_{\min}=0.01$,
and refinement strength $\eta=1.0$ at inference.

\paragraph{Evaluation and statistical testing.}
Metrics are computed after converting each predicted normalized residual back
to physical SSR units. We report MAE and RMSE in $\mathrm{W\,m^{-2}}$ for
irradiance error, SSIM for structural agreement, and LPIPS
and FID for perceptual and distributional fidelity. Lower values are better for MAE, RMSE, LPIPS, and FID; higher values are better for SSIM. For scene-complexity analysis in the main paper, gain is defined as baseline MAE minus SolarFlowRefiner MAE, so positive values indicate
lower error for SolarFlowRefiner.

We additionally test whether the SolarFlowRefiner MAE improvements are
consistent across matched held-out samples using two-sided paired Wilcoxon
signed-rank tests on the same 8,994 test samples. The tests use paired
per-sample MAE differences rather than aggregate table values.

\begin{table}[h]
\centering
\caption{Paired Wilcoxon signed-rank tests comparing per-sample MAE between each learned baseline and SolarFlowRefiner on the 8,994 held-out test samples.}
\label{tab:app_wilcoxon}
\tiny
\begin{tabular}{lccc}
\toprule
Baseline & Mean MAE gain & Lower-error samples (\%) & $p$-value \\
\midrule
ResShift & 2.97 & 73.5 & $<10^{-300}$ \\
FlowMatch & 4.24 & 66.4 & $<10^{-300}$ \\
FlowRefiner-ODE & 4.11 & 63.6 & $3.23\times10^{-304}$ \\
FlowRefiner-PDE & 2.63 & 69.0 & $<10^{-300}$ \\
\bottomrule
\end{tabular}
\end{table}

All four tests reject equality of paired MAE distributions at conventional
significance levels. Because multiple hours from the same site can still be
correlated, these paired tests should be interpreted together with the
day-blocked split and site-wise spatial analysis.

% If so, uncomment the following line after finalizing the submission.

\end{document}